\documentclass{article}

\usepackage[preprint]{neurips_2025}

\usepackage[utf8]{inputenc} 
\usepackage[T1]{fontenc}    
\usepackage{hyperref}       
\usepackage{url}            
\usepackage{booktabs}       
\usepackage{amsfonts}       
\usepackage{nicefrac}       
\usepackage{microtype}      
\usepackage{xcolor}         

\usepackage{amsmath}
\usepackage{enumitem}
\usepackage{graphicx}
\usepackage{wrapfig}

\title{Uni-ViGU: Towards Unified Video Generation and Understanding via A Diffusion-Based Video Generator [Preview]}

\author{
Luozheng Qin$^{1,}$\thanks{Equal contribution}\quad
Jia Gong$^{1,*}$\thanks{Corresponding author}\quad
Qian Qiao$^{3,*}$\quad\\
\textbf{Tianjiao Li}$^{4}$\quad
\textbf{Li Xu}$^{4}$\quad
\textbf{Haoyu Pan}$^{1,*}$\quad
\textbf{Chao Qu}$^{2,1}$\quad
\textbf{Zhiyu Tan}$^{2,1}$\quad
\textbf{Hao Li}$^{2,1\dagger}$\\
$^{1}$Shanghai Academy of AI for Science\quad
$^{2}$Fudan University\quad\\
$^{3}$Independent Researcher\quad
$^{4}$Singapore University of Technology and Design\\
}

\begin{document}

\maketitle

\begin{abstract}
Unified multimodal models integrating visual understanding and generation face a fundamental challenge: visual generation incurs substantially higher computational costs than understanding, particularly for video. This imbalance motivates us to invert the conventional paradigm: rather than extending understanding-centric MLLMs to support generation, we propose Uni-ViGU, a framework that unifies video generation and understanding by extending a video generator as the foundation. We introduce a unified flow method that performs continuous flow matching for video and discrete flow matching for text within a single process, enabling coherent multimodal generation. We further propose a modality-driven MoE-based framework that augments Transformer blocks with lightweight layers for text generation while preserving generative priors. To repurpose generation knowledge for understanding, we design a bidirectional training mechanism with two stages: Knowledge Recall reconstructs input prompts to leverage learned text-video correspondences, while Capability Refinement fine-tunes on detailed captions to establish discriminative shared representations. Experiments demonstrate that Uni-ViGU achieves competitive performance on both video generation and understanding, validating generation-centric architectures as a scalable path toward unified multimodal intelligence. Project
Page and Code: https://fr0zencrane.github.io/uni-vigu-page/.
\end{abstract}

\section{Introduction}
\label{sec:introduction}
Unified multimodal models~\cite{wu2025janus, xie2024show, deng2025bagel, qin2025unicot} aim to integrate image (and video) understanding and generation within a single framework. Such unification holds great promise for streamlining model design~\cite{wang2022ofa,bao2023one,luunified}, enabling representation sharing across tasks~\cite{team2023gemini,wang2023image}, and scaling toward general-purpose visual intelligence~\cite{team2024chameleon}. Motivated by these advantages, a growing body of research has explored diverse architectural paradigms in search of an effective and principled unified solution.

To achieve this unification, early approaches~\cite{wu2025janus,chen2025janus,team2025nextstep} formulate image generation as an autoregressive sequence prediction problem within multimodal large language models (MLLMs). However, such causal generation often yields limited visual fidelity~\cite{ControlAR,ak2020incorporating}. 
To overcome this limitation, subsequent methods~\cite{xie2024show,tong2025metamorph,chen2025blip3,yang2026omni,pan2025transfer} reformulate MLLMs~\cite{hui2024qwen2,llama3modelcard,li2023textbooks} within a diffusion framework. Specifically, they freeze a pretrained MLLM and introduce a set of learnable query tokens, which serve as the interface to enable diffusion-based image or video generation.
While this strategy substantially improves generation quality, the decoupled training paradigm prevents generation objectives from directly benefiting visual understanding. 
To bridge this gap, recent work~\cite{deng2025emerging,liang2025mixtureoftransformers} proposes a dual-tower unified framework to couples understanding and generation. Specifically, in addition to an MLLM for visual understanding, it trains a duplicated MLLM as a generator, and integrates the two branches via cross-attention to achieve more tightly unified multi-modal models.

Despite these advances, effectively unifying visual understanding and generation remains challenging. A fundamental obstacle lies in the \textit{substantially higher computational cost of visual generation compared to visual understanding}. In particular, diffusion-based generation models rely on amounts of iterative denoising steps, resulting in significant token consumption and computational overhead. For instance, generating a single image may require thousands of tokens (e.g., 4096 tokens with 50 denoising iterations for a $1024\times1024$ image in FLUX.1 dev\cite{flux1dev2024}). When extended to video generation, this cost increases dramatically due to temporal expansion across frames, easily reaching {millions of tokens per sample} (e.g., 73,920 tokens with 40-50 repeated steps for a 5-second 720P video in Wan2.2\cite{wan2025wan}). Such prohibitive complexity substantially increases both training and inference costs, making it difficult to scale MLLMs to support {high-quality visual generation alongside visual understanding}, particularly when extending from static images to videos.

These prohibitive costs motivate a natural inversion of perspective:
\textbf{\textit{rather than asking how understanding-centric MLLMs can be extended to support generation, can generation-centric models themselves serve as the architectural foundation for unified multi-modal intelligence?}}
From a developmental standpoint, humans acquire visual perceptual abilities prior to developing the linguistic capacity to articulate what they perceive~\cite{o2026infants,yeung2009learning,wang2021infant}. This asymmetry suggests that strong generative visual priors may constitute a more natural foundation for unifying understanding and generation.
From a computational perspective, the imbalance is equally pronounced: synthesizing even a short video can require processing millions of tokens, whereas generating long-form text typically involves orders of magnitude fewer tokens. This disparity underscores that visual generation dominates the computational cost in multimodal systems, further suggesting that extending a video generator toward unified intelligence may offer a more scalable and principled path.

Motivated by these observations, we propose \textbf{Uni-ViGU}, a diffusion-based framework that \textbf{uni}fies \textbf{vi}deo \textbf{g}eneration and \textbf{u}nderstanding by extending a video generator as the unified foundation.
Specifically, to achieve this paradigm, the first challenge is how to enable a video generator to produce coherent text. Modern video generation models~\cite{opensora,wan2025wan,ho2022video,yangcogvideox} are typically built on diffusion frameworks that iteratively denoise Gaussian noise into video samples, whereas text generation~\cite{hui2024qwen2,touvron2023llama,guo2025deepseek} mainly relies on autoregressive token prediction. This architectural mismatch makes direct integration non-trivial. Drawing inspiration from recent advances in diffusion-based language modeling~\cite{gongscaling,gongdiffuseq,lipman2024flowmatchingguidecode}, we introduce a \textit{unified flow method} that simultaneously performs \textit{continuous flow matching} for video generation and \textit{discrete flow matching} for text generation. This formulation enables a single generative process to produce both modalities, preserving the strengths of diffusion-based video synthesis while extending the model to text generation.

Building on this architecture, we ask a deeper question: \textit{can the knowledge embedded in video generators be repurposed for video understanding?} Our key intuition is that if generation learns a mapping from text to video, then understanding can be viewed as the reverse process. Exploiting this duality may substantially reduce the difficulty of learning video understanding.
To this end, we propose a \textit{modality-driven MoE-based unified framework} that augments each Transformer block of the original video generator with lightweight linear layers for text generation. This design preserves strong generative priors in the pretrained KV layers, while disentangling video and text generation via different FFN layers to maximize performance.
We further introduce a \textit{bidirectional training mechanism} with two stages: \textit{Knowledge Recall} and \textit{Capability Refinement}. In the Knowledge Recall stage, the model reconstructs input prompts under heavy dropout, encouraging reuse of learned text–video correspondences. Since generation typically relies on coarse prompts whereas understanding demands fine-grained semantics, we then perform Capability Refinement by fine-tuning the model to produce detailed video captions, enforcing a more discriminative shared space and improving semantically grounded video understanding.
\section{Preliminary}
\label{sec:preliminary}

In this work, we build our unified model on top of WAN2.1~\cite{wan2025wan}, a state-of-the-art and efficient text-to-video generator. For completeness, we briefly summarize its core design to contextualize our framework. Notably, since WAN2.1 adopts a standard latent diffusion paradigm, our method can naturally be extended to other video generation models following similar architectures.

\paragraph{Process of Video Generation.}
Modern video generators~\cite{wan2025wan,ho2022video,yangcogvideox} (including WAN2.1) predominantly rely on diffusion processes, synthesizing videos by iteratively denoising Gaussian noise into high‑quality outputs. To improve efficiency, this procedure is typically performed in the latent space via a variational autoencoder (VAE), which compresses pixel-level information into a compact representation, thereby reducing the computational cost of modeling high-dimensional video data.

\underline{Training.} Given a video $x$, it is first encoded into a latent representation $z_1 = \mathcal{E}(x)$ via the VAE encoder $\mathcal{E}$. The model then learns the diffusion process by constructing a sequence of intermediate latents $\{z_t\}_{t=0}^1$ between $z_1$ and a Gaussian noise sample $z_0 \sim \mathcal{N}(0, I)$. Specifically, WAN2.1 follows flow matching~\cite{lipmanflow,esser2024scaling} (a variant of diffusion methods) and defines $z_t$ through linear interpolation:
\begin{equation}
z_t = (1 - t) z_0 + t z_1, \quad t \sim \mathcal{U}(0,1).
\end{equation}
This formulation defines a transport path from noise $z_0$ to data $z_1$ with a constant velocity $u = z_1 - z_0$.

A neural network $v_\theta$ is then trained to predict this target velocity $u$, conditioned on the text prompt $c$, the intermediate latent $z_t$, and the time step $t$:
\begin{equation}
\mathcal{L}_{\mathrm{FM}} =
\mathbb{E}_{x,\, z_0,\, t,\, c}
\left[
\left\| 
v_\theta(z_t, t, c) - (z_1 - z_0)
\right\|_2^2
\right],
\quad z_1 = \mathcal{E}(x).
\end{equation}

\underline{Inference.}
At test time, generation begins from Gaussian noise $z_0 \sim \mathcal{N}(0, I)$. The model then iteratively denoises the latent using the learned velocity field:
\begin{equation}
z_{t+\Delta t} = z_t + \Delta t\, v_\theta(z_t, t, c),
\quad z_0 \sim \mathcal{N}(0, I).
\end{equation}

By recurrently applying this update from $t = 0$ to $t = 1$, the process produces the final latent $z_1$, which is decoded into the output video $\hat{x} = \mathcal{D}(z_1)$ via the decoder of VAE.

\paragraph{Architecture of Video Generator}
The video generation framework comprises three key components: a VAE for compression, a text encoder for conditioning, and a diffusion model for generation.

\underline{VAE.} WAN2.1 adopts a causal VAE that decomposes a video into a set of chunks and then uses 3D convolution layers to sequentially encode each chunk into a compact latent representation. In general, it maps a video $x$ of shape $(T \times H \times W)$ to a sequence of latent features $\{ z_i \}_{i=1}^L$, where the length $T$ equals to $(1 + T/4) \times H/16 \times W/16$, the dimension of $z_i$ equals to 1536.

\underline{Text Encoder.}
WAN2.1 adopts a pretrained text encoder (umT5~\cite{chung2023unimax}) transforms input text $y$ into embeddings $c = \text{Enc}(y)$, whose dimension equals to 4096.

\begin{wrapfigure}{r}{0.55\textwidth}
  \centering
  \includegraphics[width=0.55\textwidth]{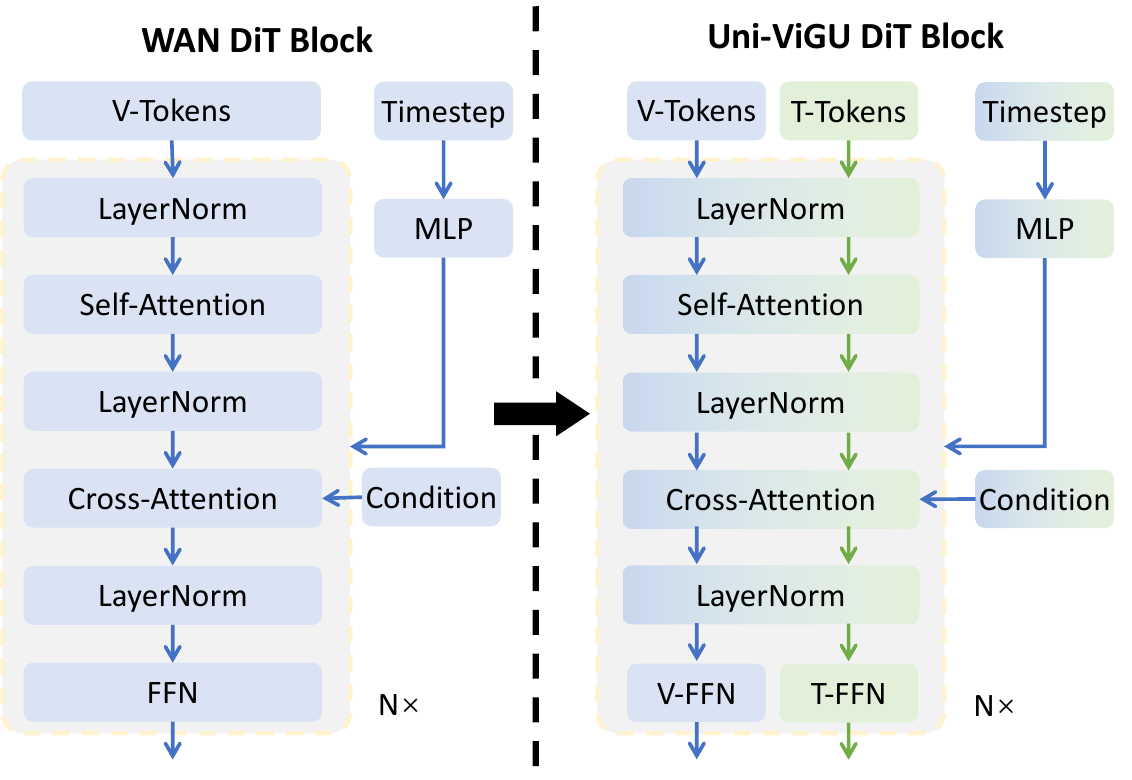}
  \caption{The DiT architecture of WAN and Uni-ViGU}
  \vspace{2pt}
  \label{fig:dit_block}
\end{wrapfigure}
\underline{Diffusion Backbone.}  
WAN2.1 adopts Diffusion Transformers (DiT) as its diffusion model. DiT is composed of multiple transformer blocks, each containing a self-attention layer, a cross-attention layer, and an FNN layer. As shown in the left of Figure~\ref{fig:dit_block}, given an input $x$ with initial hidden state $h_0$, each block updates the hidden representation as follows:
\begin{equation}
\tilde{h}_{k} = \text{CrossAttn}\!\big(\text{SelfAttn}(h_k),\; c\big),
\end{equation}
\begin{equation}
h_{k+1} = \text{FNN}\!\big(\tilde{h}_{k}\big),
\end{equation}
where $c$ denotes the text conditioning used as the key–value pair in the cross-attention layer. The self-attention module captures spatial and temporal dependencies within the video features, while cross-attention injects semantic information from the text prompt.
\section{Method}
\label{sec:method}
Having reviewed the video generation process, we now introduce our generation-based unified framework that extends the video generator to support both video understanding and video generation within a single model. Section~\ref{subsec:unified_diffusion} describes how we unify text and video generation via a single uni-flow process. Section~\ref{subsec:model_design} presents our modality-driven MoE architecture for effectively learning this uni-flow process. Finally, Section~\ref{subsec:training_inference} details our training and inference procedures.

\begin{figure}[t]
\includegraphics[width=\linewidth]{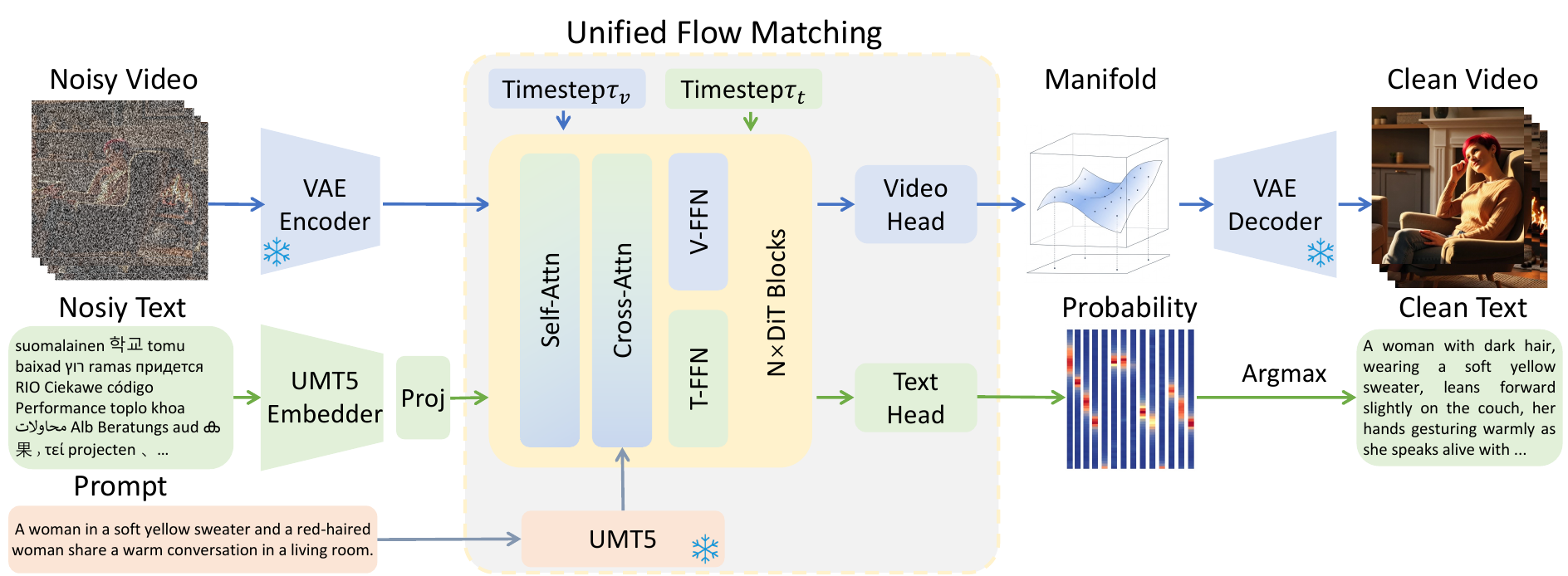}
\vspace{-6mm}
\caption{Overview of Uni-ViGU framework. We formulate unified multimodal generation via a uni-flow process, where video generation is modeled with continuous flow matching in the latent space, while text generation is modeled with discrete flow matching over token embeddings. Both modalities are jointly learned within a single Transformer backbone: self- and cross-attention layers are shared to enable cross-modal interaction and alignment, whereas modality-specific FFN branches are employed to capture domain-specific knowledge for video and text, respectively. This design allows the model to unify generation and understanding while preserving modality-specialized capacity.}
\vspace{-3mm}
\label{fig:overview}
\end{figure}

\subsection{Unifying Text and Video Generation via Uni-Flow}
\label{subsec:unified_diffusion}

Our central goal is to endow a video generator with language reasoning and comprehension abilities. The key challenge lies in the fundamental mismatch between modalities: video exists in a continuous latent space naturally suited for diffusion, while text consists of discrete tokens traditionally generated via autoregressive prediction. To bridge this gap, as shown in Figure.~\ref{fig:overview}, we propose a uni-flow process that performs \emph{continuous flow matching} for video and \emph{discrete flow matching} for text within a single generative process. We first describe each formulation independently, then show how they can be elegantly unified.

\paragraph{Continuous Flow Matching for Video.}
As introduced in the Section.~\ref{sec:preliminary}, video generation operates in the continuous latent space of a VAE. Given a video encoded as $z_{v,1} = \mathcal{E}(x)$ and noise $z_{v,0} \sim \mathcal{N}(0, I)$, flow matching constructs a transport path via linear interpolation:
\begin{equation}
z_{v,\tau} = (1 - \tau) z_{v,0} + \tau z_{v,1}, \quad \tau \sim \mathcal{U}(0,1).
\end{equation}
The model learns to predict the velocity $u_v = z_{v,1} - z_{v,0}$, which defines a continuous vector field transporting Gaussian noise to the data distribution. At inference, an ODE solver integrates this field to generate videos. This continuous formulation is well-established and forms the backbone of modern video generators such as WAN2.1.

\paragraph{Discrete Flow Matching for Text.}
Text generation presents a distinct challenge: the output space is inherently discrete (vocabulary tokens), yet we seek to model it within the same diffusion framework. Recent advances in discrete diffusion~\citep{austin2021structured,li2022diffusion} demonstrate that flow matching can be extended to discrete data by operating over token embeddings. Specifically, let $y = (y_1, \ldots, y_N)$ denote a text sequence of $N$ tokens. We map each token to a continuous embedding via a learnable matrix $E \in \mathbb{R}^{V \times d}$, yielding the text latent $z_{t,1} = (E_{y_1}, \ldots, E_{y_N}) \in \mathbb{R}^{N \times d}$. We then apply flow matching in this embedding space:
\begin{equation}
z_{t,\tau} = (1 - \tau) z_{t,0} + \tau z_{t,1}, \quad z_{t,0} \sim \mathcal{N}(0, I), \quad \tau \sim \mathcal{U}(0,1).
\end{equation}
The model predicts the velocity $u_t = z_{t,1} - z_{t,0}$, learning to transport noise to the manifold of valid token embeddings. At inference, after integrating to obtain $z_{t,1}$, we decode to discrete tokens by computing similarity to the embedding matrix:
\begin{equation}
\hat{y}_i = \arg\max_{v \in \mathcal{V}} \; z_{t,1}^{(i)} \cdot E_v,
\end{equation}
where $z_{t,1}^{(i)}$ denotes the $i$-th token embedding and $\mathcal{V}$ is the vocabulary. 
To facilitate stable training, we further encourage the token embeddings in $\mathcal{V}$ to align with the continuous video manifold.

\paragraph{Unified Flow Matching.}
Building upon above formulation, we model the joint video-text distribution $q_{\mathrm{data}}(z_v, z_t \mid c)$ by simultaneously constructing flow matching paths for both modalities:
\begin{equation}
z_{v,\tau_v} = (1 - \tau_v) z_{v,0} + \tau_v z_{v,1}, \quad z_{t,\tau_t} = (1 - \tau_t) z_{t,0} + \tau_t z_{t,1},
\end{equation}
where crucially, $\tau_v, \tau_t \sim \mathcal{U}(0,1)$ are \emph{independently sampled}. This independence is a key design: it allows the two modalities to progress through different stages of their respective denoising processes, enabling the model to learn cross-modal dependencies across all combinations of noise levels.

With this design, a unified model $f_\theta$, condtioned on the prompt $c$, jointly predicts both velocity fields:
\begin{equation}
[\hat{v}_v, \hat{v}_t] = f_\theta(z_{v,\tau_v}, z_{t,\tau_t}, \tau_v, \tau_t, c),
\end{equation}
optimized via the combined objective:
\begin{equation}
\mathcal{L}_{\mathrm{UFM}} = \mathbb{E} \left[ \lambda_v \| \hat{v}_v - u_v \|_2^2 + \lambda_t \| \hat{v}_t - u_t \|_2^2 \right],
\end{equation}
where $\lambda_v = 1.0$ and $\lambda_t = |z_v|/|z_t|$ normalize contributions by token count. This unified framework naturally accommodates both tasks: setting $\tau_v = 1, \tau_t = 0$ enables video understanding (clean video, noisy text), while $\tau_v = 0, \tau_t = 1$ enables video generation (noisy video, clean text).

\subsection{Modality-Driven Mixture-of-Experts Architecture}
\label{subsec:model_design}

Having established the unified flow matching framework, we now instantiate it using pretrained video generators. Our key observation is that video DiTs implicitly encode rich visual–semantic correspondences through large-scale text-to-video pretraining. We hypothesize that this learned alignment can be effectively repurposed for video-to-text generation with minimal architectural modifications. The central question then becomes: \textit{where does such transferable cross-modal knowledge reside within the network?}

To answer this, we revisit the functional roles of each component in a video DiT block. As discussed in Section.~\ref{sec:preliminary}, a standard block consists of self-attention, cross-attention, and feed-forward network (FFN) layers, each serving distinct computational purposes. Given token representations $h \in \mathbb{R}^{n \times d}$, attention computes:
\begin{equation}
\text{Attention}(h) = \text{softmax}\left(\frac{QK^\top}{\sqrt{d}}\right)V, \quad Q, K, V = hW_Q, hW_K, hW_V,
\end{equation}
where each token aggregates information from all others, thereby capturing \textit{relational} structure across the sequence. In contrast, FFN layers apply position-wise transformations:
\begin{equation}
\text{FFN}(h_i) = W_2 \cdot \sigma(W_1 h_i + b_1) + b_2,
\end{equation}
which process each token independently and thus primarily encode \textit{domain-specific} knowledge. This functional decomposition reveals a natural division of labor: cross-modal alignment—being inherently relational—is predominantly captured by attention layers, whereas modality-specific generation patterns are governed by FFN layers.

This analysis directly motivates our architectural design principle: \textbf{share attention to preserve cross-modal alignment, while separating FFN layers to accommodate modality-specific generation}. Concretely, as shown in the right of Figure.~\ref{fig:dit_block}, attention operates over the concatenation of video and text tokens:
\begin{equation}
[\tilde{h}_v; \tilde{h}_t] = \text{Attention}([h_v; h_t]),
\end{equation}
enabling bidirectional cross-modal interaction through shared attention patterns. The resulting representations are then routed to modality-specific experts:
\begin{equation}
h'_m = \text{FFN}_m(\tilde{h}_m), \quad m \in \{v, t\},
\end{equation}
where routing is deterministic based on modality identity. This can be viewed as a structured Mixture-of-Experts (MoE), but unlike conventional MoE architectures that rely on learned gating mechanisms~\citep{jacobs1991adaptive,dai2024deepseekmoe}, our design enforces explicit modality separation while fully preserving the shared relational reasoning learned during pretraining.

The initialization strategy follows naturally from our goal of knowledge transfer. The video expert $\text{FFN}_v$ retains pretrained weights to preserve generative priors, whereas the text expert $\text{FFN}_t$ is newly initialized to support text generation. This asymmetric initialization, combined with the shared-attention design, yields three practical benefits: (1) attention parameters are fully shared across modalities, maximizing knowledge reuse; (2) FFN duplication incurs only minimal parameter overhead; and (3) the preserved pretrained components facilitate rapid convergence during training.

\subsection{Training and Inference}
\label{subsec:training_inference}

\paragraph{Bidirectional Training.} 
To enable effective knowledge reutilization and skill development, we propose a bidirectional two-stage training framework: the Knowledge Recall Stage and the Capability Refinement Stage.
Specifically, we first initialize our model with a pretrained video generator (Wan2.1). We then train the model to learn video-text mappings during the Knowledge Recall Stage, followed by developing video understanding capabilities during the Capability Refinement Stage.

\underline{Stage~1: Knowledge Recall.}
In this stage, the target text $z_{t,1}$ is set identical to the conditioning prompt $c$ itself. Since the video generator has been pretrained to learn the mapping from prompt $c$ to video $v$, the model should readily learn the reverse mapping from video $v$ to the target text (which equals $c$), as this leverages the same correspondence already encoded in its parameters.

However, if the conditioning prompt $c$ remains available during training, the model can trivially copy $c$ to predict $z_{t,1}$ without actually extracting information from the video. To eliminate this shortcut, we apply \textit{condition dropout}, dropping $c$ with probability $p$. This forces the model to recover the text from the co-noised video latent $z_{v,\tau_v}$, compelling it to leverage its pretrained text-to-video correspondences in the reverse direction. This stage serves as an efficient warm-up that rapidly adapts the model's generative prior from single modality to the joint video-text unified flow matching formulation, establishing basic cross-modal alignment with minimal training cost.

Furthermore, a substantial imbalance exists during model training: video latents contribute approximately 30K tokens, while the text sequence comprises only 256 tokens. Moreover, video generation has already been well modeled during pretraining, whereas text generation represents a novel task for the model. These observations jointly suggest that the text generation branch should receive greater optimization emphasis. Accordingly, we set $\lambda_v = 1.0$ for video and $\lambda_t = |z_v| / |z_t|$ for text, where $|z_v|$ and $|z_t|$ denote the number of video and text tokens, respectively. This token-count normalization ensures that each modality receives balanced per-token supervision.

\underline{Stage~2: Capability Refinement.}
While Stage~1 activates cross-modal knowledge transfer, the conditioning prompt $c$ is typically a brief, coarse description, insufficient for fine-grained video understanding. In this stage, we replace the target text with \textit{detailed video captions} that provide rich, semantically precise descriptions of the visual content. 
Since these detailed captions contain substantially more information than the conditioning prompt, the text generation branch can no longer rely on the cross-attention features from $c$ alone. Instead, it must actively attend to the video latent $z_{v,\tau_v}$ through the shared self-attention mechanism to recover fine-grained visual details, object attributes, spatial relationships, temporal dynamics, that are absent from the brief prompt but present in the video. 
This forces the model to develop genuine video comprehension capabilities, fully exploiting the bidirectional cross-modal interaction enabled by our unified framework and yielding deeply aligned multimodal representations.

\paragraph{Inference.}
For {video understanding}, we fix $z_{v,\tau_v} = z_{v,1}$ (clean video) and integrate the text flow from noise:
\begin{equation}
z_{t,\tau+\Delta\tau} = z_{t,\tau} + \Delta\tau \cdot \hat{v}_t, \quad z_{t,0} \sim \mathcal{N}(0, I).
\end{equation}
The final $z_{t,1}$ is decoded to tokens via embedding lookup.

For {video generation}, we fix $z_{t,\tau_t} = z_{t,1}$ (embedded prompt) and integrate the video flow:
\begin{equation}
z_{v,\tau+\Delta\tau} = z_{v,\tau} + \Delta\tau \cdot \hat{v}_v, \quad z_{v,0} \sim \mathcal{N}(0, I).
\end{equation}
The final $z_{v,1}$ is decoded to pixels via the VAE.

This symmetric procedure realizes bidirectional video-text mapping within a single unified model.

As for the \textbf{joint video-text generation}, both modalities are initialized from noise and denoised simultaneously. Specifically, we sample $z_{v,0} \sim \mathcal{N}(0, I)$ and $z_{t,0} \sim \mathcal{N}(0, I)$, and integrate both flows in parallel:
\begin{equation}
z_{v,\tau+\Delta\tau} = z_{v,\tau} + \Delta\tau \cdot \hat{v}_v, \quad
z_{t,\tau+\Delta\tau} = z_{t,\tau} + \Delta\tau \cdot \hat{v}_t,
\end{equation}
where at each step the two streams are coupled through shared self-attention: the partially denoised text latents $z_{t,\tau}$ provide progressively refined semantic guidance for video denoising, while the emerging video latents $z_{v,\tau}$ supply visual context that steers text generation. As the flow progresses from $\tau=0$ to $\tau=1$, the two modalities co-evolve and mutually sharpen each other, yielding a coherent and deeply aligned video-text pair upon convergence. 
\section{Experiment}

\subsection{Implementation Details}

\paragraph{Dataset.}
To model the joint video-text distribution, Uni-ViGU is trained on meticulously curated video-text pairs following the two-stage bidirectional training framework described in Section~\ref{subsec:training_inference}.

In \textbf{Stage~1 (Knowledge Recall)}, the model is trained on 10K video-prompt pairs. As described in Section~\ref{subsec:training_inference}, the target text $z_{t,1}$ is set identical to the conditioning prompt $c$, and condition dropout is applied to prevent trivial copying. 
In \textbf{Stage~2 (Capability Refinement)}, the model is further fine-tuned on an additional 10K video-prompt-detailed caption triples. 
It is conditioned on the brief prompt $c$ while its text generation target is replaced with a detailed caption that provides rich, semantically precise descriptions aligned with the video content.

The training data is constructed as follows. We first prepare a set of conditioning prompts and use state-of-the-art video generators to synthesize videos from these prompts. 
A LLM is then employed to jointly comprehend each video-prompt pair and produce a detailed caption that faithfully covers every aspect described by the input prompt while substantially enriching the level of detail. We enforce token-length constraints on the paired data: conditioning prompts are restricted to 0-128 tokens and detailed captions to 128-256 tokens. 
This length separation ensures that the detailed caption cannot be trivially inferred from the conditioning prompt alone, establishing a solid foundation for training convergence and compelling the model to develop genuine video comprehension capabilities through the shared attention mechanism (Section~\ref{subsec:model_design}).

\paragraph{Training Setup.}
We initialize our model from the pretrained Wan2.1~\cite{wan2025wan} video generator, as described in Section~\ref{sec:preliminary}. 
The video expert $\text{FFN}_v$ retains pretrained weights, while the text expert $\text{FFN}_t$ is newly initialized (Section~\ref{subsec:model_design}). 
Stage~1 is trained for 40K steps with a learning rate of $2 \times 10^{-4}$, and Stage~2 is trained for 60K steps with a learning rate of $5 \times 10^{-5}$. 
Both stages use the Adam optimizer with $\beta_1 = 0.90$ and $\beta_2 = 0.95$. The loss weights follow the token-count normalization scheme described in Section~\ref{subsec:training_inference}, with $\lambda_v = 1.0$ and $\lambda_t = |z_v|/|z_t|$. 
The total training cost for Uni-ViGU is 16 H800 GPUs within one week.

\subsection{Results on Joint Video-Text Generation}
\begin{figure}[t]
\includegraphics[width=\linewidth]{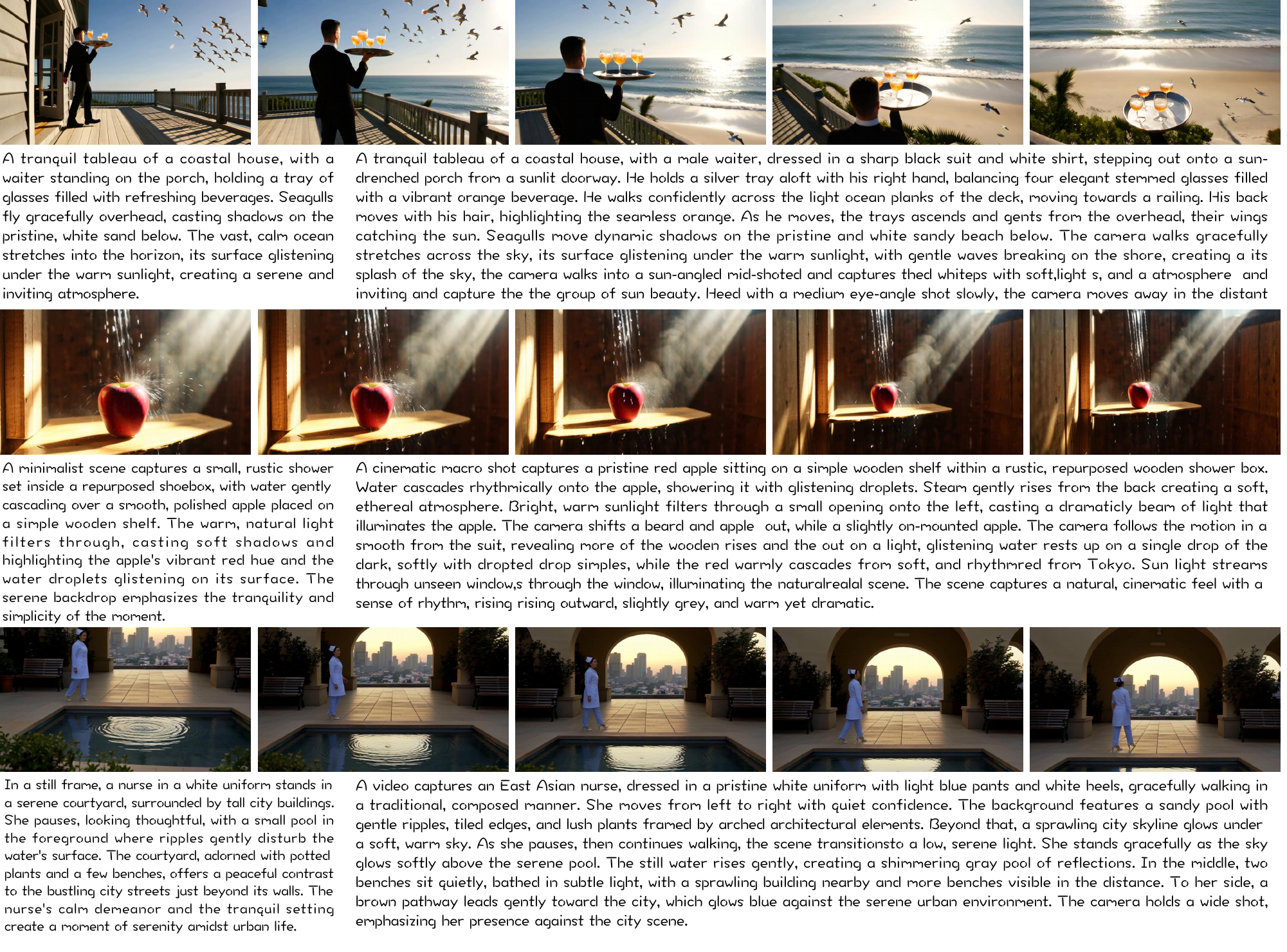}
\vspace{-6mm}
\caption{Qualitative Results on Video-Text Joint Generation}
\vspace{-3mm}
\label{fig:joint_cases}
\end{figure}

We mainly evaluate the joint generation capability of Uni-ViGU, where both video and text are simultaneously denoised from Gaussian noise. 
As described in Section~\ref{subsec:training_inference} and illustrated in Figure~\ref{fig:joint_cases}, the two modalities co-evolve through shared attention module: the progressively denoised text latents provide increasingly precise semantic guidance for video generation, while the emerging video latents supply rich visual context that steers text generation. 
Through this mutual refinement process, the model produces high-quality videos paired with detailed captions that are substantially more descriptive and faithful to the visual content than the original conditioning prompts.

\section{Related Work}

\paragraph{Unified Multimodal Understanding and Generation}
Integrating visual understanding and generation within a single model has gained significant attention. Early approaches cast image generation as autoregressive prediction in multimodal LLMs (MLLMs), mapping visual signals to discrete tokens via a shared vocabulary \citep{wu2025janus,team2024chameleon,xie2024show,jiao2025unitoken}. However, discrete tokenization inevitably sacrifices high-frequency visual details. Subsequent methods \citep{tong2025metamorph,chen2025blip3,pan2025transfer} preserve fidelity by retrofitting MLLMs with continuous diffusion modules, though their decoupled training prevents generation objectives from benefiting understanding. Recent dual-tower frameworks \citep{deng2025emerging,liang2025mixtureoftransformers} achieve tighter integration via cross-attention between understanding and generation branches.

Despite this progress, a fundamental bottleneck remains: visual generation is far more computationally expensive than understanding, especially for video, where iterative denoising over millions of tokens makes extending understanding-centric MLLMs prohibitive. We invert this paradigm, rather than augmenting MLLMs for generation, we extend video generators to support understanding, leveraging their rich spatiotemporal priors as a more scalable foundation. The most related works are \cite{bao2023one,li2026omni}, which unify text and image generation via diffusion. However, a key distinction exists: our approach re-utilizes pretrained generation knowledge from diffusion-based video models rather than training generation from scratch. This enables us to achieve unified video understanding and generation with substantially lower computational overhead.

\paragraph{Video Generation Models}

Video generation has undergone a paradigm shift from early 3D U-Net architectures \citep{blattmann2023align} to Diffusion Transformers (DiTs) \citep{peebles2023scalable,brooks2024video}, which offer superior scalability for modeling complex temporal dynamics. Modern systems such as Wan \citep{wan2025wan}, CogVideoX \citep{yangcogvideox}, and OpenSora \citep{opensora} demonstrate that DiT-based architectures can synthesize high-fidelity, long-form videos. Continuous flow matching \citep{lipman2022flow,liu2022flow} has emerged as an efficient training formulation, enabling faster convergence than standard diffusion objectives.

Crucially, these video generators learn rich text-to-video correspondences through large-scale pretraining: they must understand textual descriptions sufficiently well to synthesize semantically aligned visual content. However, this implicit visual-semantic knowledge remains confined to the generation pathway and is rarely exploited for explicit language-level comprehension. Our framework repurposes these learned correspondences bidirectionally, if generation learns mappings from text to video, understanding can leverage the reverse mapping, substantially reducing the difficulty of learning video comprehension from scratch.

\paragraph{Diffusion-Based Language Modeling}

Autoregressive next-token prediction dominates text generation, yet its strict left-to-right causality conflicts with the non-causal denoising process of diffusion-based video synthesis. Recent advances demonstrate that diffusion frameworks can effectively model discrete text. Discrete state-space diffusion \citep{lou2023discrete}, masked diffusion language models \citep{sahoo2024simple}, and discrete flow matching approaches \citep{gongscaling,lipman2024flowmatchingguidecode} show that non-autoregressive language modeling scales competitively with autoregressive baselines. LLaDA \citep{nie2025large} and related methods further establish that diffusion-based text generation can achieve strong performance across diverse benchmarks.

However, these diffusion language models are typically developed in isolation from visual synthesis. Our unified flow formulation bridges this gap by performing continuous flow matching for video and discrete flow matching for text within a single generative process. This alignment under a shared objective eliminates the architectural fragmentation caused by mismatched generation paradigms, enabling coherent joint optimization of both modalities.

\section{Conclusion}
In this work, we presented Uni-ViGU, a unified framework that extends pretrained video generators to support both video generation and understanding. Our key insight is that the rich visual-semantic correspondences learned during text-to-video pretraining can be repurposed for video understanding by treating it as the inverse of generation. We introduced a uni-flow formulation that unifies continuous flow matching for video and discrete flow matching for text, a modality-driven MoE architecture that shares attention while separating FFN layers, and a bidirectional training mechanism that progressively activates cross-modal knowledge transfer. Our generation-centric paradigm offers a principled and scalable alternative to understanding-centric approaches, opening promising directions for unified multi-modal intelligence.

\clearpage
%
%
\bibliographystyle{unsrt}
\bibliography{egbib}

\end{document}